\def\year{2020}\relax
\documentclass[letterpaper]{article} 
\usepackage{aaai20}  
\usepackage{times}  
\usepackage{helvet} 
\usepackage{courier}  
\usepackage[hyphens]{url}  
\usepackage{graphicx} 
\usepackage{graphicx}  
\title{Evaluating Seq2Seq Learning Models for If-Then Program Synthesis}
\author{Dhairya Dalal and
		Byron V. Galbraith \\ 
		Talla, Inc \\
\{dhairya, byron\}@talla.com \\
2 Park Plaza, 7th Floor, Boston, MA 02116 \\
}
\begin{document}

\maketitle

\begin{abstract}
	Implementing enterprise process automation often requires significant technical expertise and engineering effort. It would be beneficial for non-technical users to be able to describe a business process in natural language and have an intelligent system generate the workflow that can be automatically executed. A building block of process automations are If-Then programs. In the consumer space, sites like IFTTT and Zapier allow users to create automations by defining If-Then programs using a graphical interface. We explore the efficacy of modeling If-Then programs as a sequence learning task. We find Seq2Seq approaches have high potential (performing strongly on the Zapier recipes) and can serve as a promising approach to more complex program synthesis challenges.  
\end{abstract}

\noindent In this paper we explore the task of program synthesis from natural language. The goal is to translate a natural language description into an executable program. We specifically consider the use case of If-Then program synthesis as it aligns closely with enterprise automation. Business processes are driven by the collection, movement, and synthesis of information produced by various information systems. Given the technical complexity and programmatic interfaces of these information systems, enterprise automation is often bottle-necked by engineering and IT effort. Non-technical business users operate at the level of business logic and high level concepts. There can be significant value in developing intelligent systems to empower non-technical business users and provide them with the ability to create automations from natural language descriptions. 

On the consumer side, services like Zapier and If Then Than That (IFTTT) allow non-technical users to easily chain together various web services and APIs to create automations. These automations are described as “recipes” and are generated by the end user using a graphical user interface. Recipes on IFTTT and Zapier follow the If-Then format. The user selects a triggering event on a web service and an ensuing action to be executed (using the same or different web service). For example, imagine a user who wishes to automatically publish breaking news on twitter. A simple If-Then recipe to achieve this could be: if a new article is posted on NY Times, publish a tweet with link to article on my Twitter.

Both IFTTT and Zapier use a point and click graphical interface which are quite intuitive. We consider the use case of generating automation recipes from natural language descriptions for two reasons. With the proliferation of chatbots and dialog systems, business users can quickly and more efficiently describe an automation using natural language. Additionally, the compactness of languages allows for the description of more complex business processes which can often be more difficult to model visually. 

In this paper, we aim to predict IFTTT and Zapier recipes from natural language descriptions. In contrast to previous approaches, we frame the task an end-to-end sequence-to-sequence learning problem. All associated code and data for this paper can be found on the author's github page: https://github.com/dhairyadalal/nmt\_if\_then

\section{If-Then Program Synthesis}
If-Then programs consist of a triggering event and an ensuing action. In this paper we explore trigger-action programs in the context of IFTTT and Zapier “recipes”. We use the IFTTT \cite{quirk-etal-2015-language} and Zapier \cite{chen2016latent} data sets. While the specific terminology for IFTTT and Zapier recipes are slightly different, the recipe formats are conceptually the same. The automation recipes consist of channel entities, functions, and function arguments. Channel entities are the specific web service providers such as Twitter or New York Times. Functions are the API services provided by the channel entities. Functions can also take in arguments. Following with previous work on the IFTTT, we did not consider argument extraction in this paper. 

\begin{figure}[t]
	\centering
	\includegraphics[width=0.9\columnwidth]{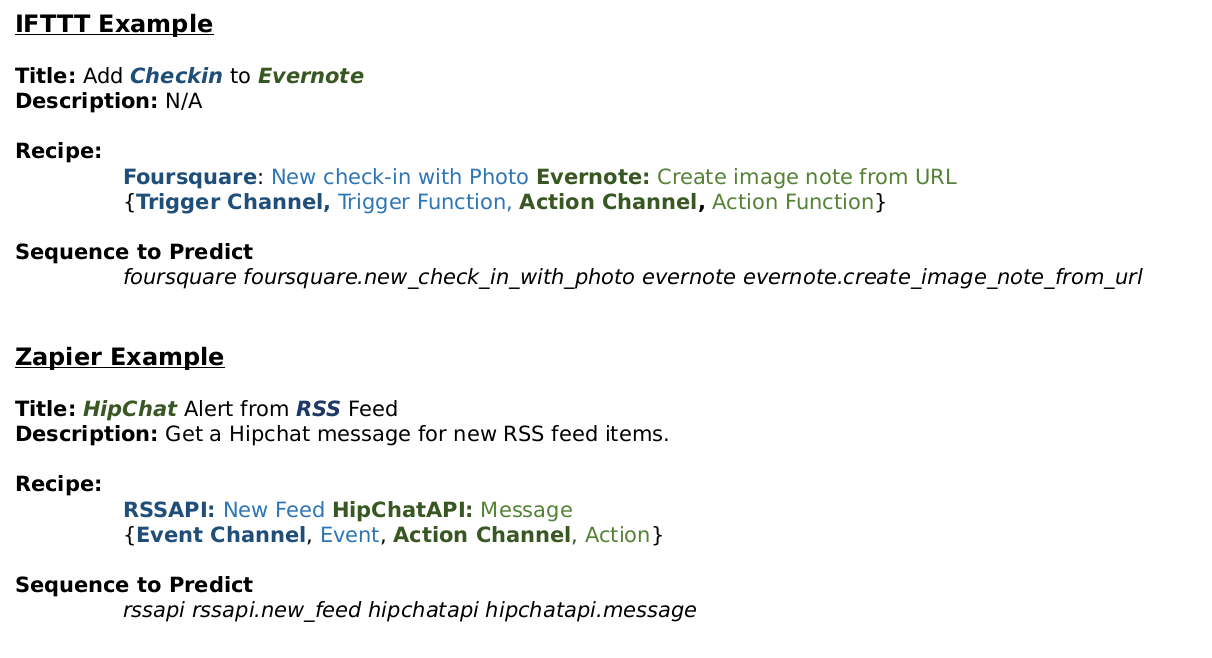} 
	\caption{Example recipes from IFTTT and Zapier}
	\label{fig1}
\end{figure}

There are many interesting challenges that arise in If-Then program synthesis. Given the examples in Figure \ref{fig1}, we can see following challenges: \\ \\
\begin{itemize}
	\item Alignment: the trigger/event channel appears in different positions across the two examples. In the IFTTT example the "If" condition is mentioned first and in the Zapier example, it appears second. 
	\item Channel/function ambiguity: in the IFTTT example, Foursquare is not explicitly mentioned in the title. We infer the act of "checkin" to be associated with Foursquare checkins. 
	\item Variable input lengths: descriptions of recipes are not the same length	 
\end{itemize}

\subsection{IFTTT}
In the IFTTT domain, an automation recipe consists of the following four elements: Trigger Channel, Trigger Function, Action Channel and Action Function. The trigger is the "If" part of the If-Then program and identifies a condition or activity that will trigger the "Then" or action that will ensue. 
 
We obtained the IFTTT dataset and relevant materials from the Microsoft Open Data Repository. \citeauthor{quirk-etal-2015-language} do not provide the scraped IFTTT recipes. Instead they provide their web scraping tool, a list of recipe urls, and the Mechanical Turk annotations for the test set. After scraping and extraction process, we have a dataset that contains the following for each recipe: recipe title, recipe description, and the formal IFTTT recipe. The quality of the recipe descriptions was highly variable and unreliable. We followed the convention of previous work and used the recipe title as our natural language description. 

Since 2015, many of the provided urls are no longer viable. Additionally, upon inspecting the dataset, we further removed examples where the title was not in English or the title contained fewer than three words. Additionally, we followed the convention set by previous work and further pruned our test set based on the Mechanical Turkers annotations. We only selected examples where three or more annotators agreed on all the extracted recipe parts. Table \ref{table2} described the size of our final dataset. In previous works the test set was slightly larger at 684 examples, however not all of those example were available during our scraping activity.

\begin{table}[t]
\begin{tabular}{lllcl}
	\hline
	Type       & Original & Removed &\% Removed	& Final   \\ \hline
	train      & 77,495   & 37,523 &48	 	of the recipe	& 39,972  \\
	validation & 5,161    & 2,615  &50 	 		& 2,546   \\
	test       & 4,294    & 3,740  &87 			& 554	  \\ \hline
\end{tabular}
\caption{IFTTT data after cleaning bad urls and bad entries.}\smallskip
\label{table2}
\end{table}

\subsection{Zapier}
Zapier recipes are conceptually identical to IFTTT recipes and consist of the following four elements: Event Channel, Event, Action Channel, and Action. Here the \textit{Event} and \textit{Action} are identical to \textit{Triggers} and \textit{Actions} in the IFTTT domain. 

We used Zapier dataset that was proposed in \cite{chen2016latent}. The authors provided a segmented train and test dataset on their github repository. The Zapier has 17,264 train examples and 4,859 test examples. We further segment the train set by randomly generating a validation set of 3,896 examples. Like the IFTTT, each example consisted of a recipe title, recipe description, and the full recipe. \citeauthor{chen2016latent} do not provide performance metrics of their model on the Zapier dataset, but they do provide their code. We were unable to get baseline metrics of the Latent Attention model on the Zapier data.

\section{Related Work}
Program synthesis from natural language is a broad research space with numerous use cases and challenges. The IFTTT recipe prediction task was first proposed by \citeauthor{quirk-etal-2015-language}. \citeauthor{quirk-etal-2015-language} initially framed the problem as a structured prediction task where channels and thier corresponding functions were predicted separately using phrasal machine translation and text-similarity retrieval methods. \citeauthor{beltagy-quirk-2016-improved} \citeyear{beltagy-quirk-2016-improved} then framed the problem as an ensemble based sequence-to-sequence prediction task. In this approach, distinct recurrent neural networks and a logistic regression classifiers were trained to predict the channels and functions as separate multi-class labels. The prediction were combined to form recipe candidates and an ensemble model was trained to select the best recipe candidate. \citeauthor{chen2016latent} \citeyear{chen2016latent} proposed an end-to-end neural architecture. While channels and functions were treated as distinct multi-class labels, the model jointly predicted all the recipe components using shared latent attention. \cite{chen2016latent} also produced the Zapier dataset but did not publish any evaluation metrics as their paper focused primarily of the IFTTT use case. Finally, \citeauthor{DBLP:journals/corr/abs-1808-06740} \citeyear{DBLP:journals/corr/abs-1808-06740} proposed an interactive reinforcement learning agent that asks a human user for clarification prior parsing the natural language description.

Outside of the If-Then program domain, various strategies and use cases have emerged. \citeauthor{Lin2017ProgramSF} \citeyear{Lin2017ProgramSF} first proposed a seq2seq approach for program synthesis using a RNN encoder-decoder to translate natural language commands into executable shell scripts. \citeauthor{krishnamurthy-etal-2017-neural} \citeyear{krishnamurthy-etal-2017-neural} developed an augmented encoder-decoder model that incorporates type-checking and entity resolutions to improve question answering on semi-structured tables. In uses cases where there is an execution environment available, reinforcement learning and weakly supervised approaches have also been proposed. 

\section{Program Synthesis as Seq2Seq Learning}
In contrast to previous approaches, we consider framing If-Then program synthesis as an end-to-end sequence learning task. The sequence-to-sequence encoder-decoder architecture was first introduced by \citeauthor{DBLP:journals/corr/SutskeverVL14} \citeyear{DBLP:journals/corr/SutskeverVL14} for machine translation. Given a sequence of inputs $\{x_1, .., x_i\}$ where $x \in Vocabulary_{source}$, the seq2seq model aimed to predict a sequence $\{y_1, .., y_i\}$ where $y \in Vocabulary_{target}$. Seq2Seq models were first used for machine translation tasks where inputs from one language was translated to another language. Given the flexibility of the architecture, the model was applied to many other domains and use cases including program synthesis.

Current approaches to If-Then program synthesis exploit the formulaic nature of the IFTTT and Zapier recipes. They represent each component of the recipe as distinct multi-class labels. A downside to this is approach is if the recipe format is expanded, additional classification heads must be added. Additionally, existing models cannot predict more complicated recipes that may contain multiple channels, complex control structures and embedded function arguments. Seq2seq models offer a higher degree of flexibly in representing complex programs as sequences. We investigated if seq2seq could be successful in If-Then program synthesis, especially around predicting the recipe in its entirety given variable inputs.

\section{Experiment}
We evaluated three different seq2seq architectures: a baseline LSTM Encoder-Decoder, OpenNMT, and basic Transformer. We intentionally selected OpenNMT and the Transformer architectures given their recent popularity and general effectiveness in neural machine translation tasks. We describe the architecture in further detail below. 

Both the IFTTT and Zapier datasets were preprocessed identically. Recipe titles were converted to lowercase. In the Zapier dataset, we found descriptions were high quality and improved all the performance of all models by 1 percent. Description were concatenated to the title using the "[SEP]" token.

The recipes were converted to a single space delimited sequence. If the channel or function consisted of multiple words, they were joined using underscores. Additionally, the channel was prefixed to the function. For example the following IFTTT recipe: \\ \\
\textbf{Trigger Channel}: NY Times \\
\textbf{Trigger Function:} New Article Posted \\
\textbf{Action Channel:} Twitter \\
\textbf{Action Function:} New Post \\ \\
was converted to the following sequence: \textit{ny\_times ny\_times.new\_article\_posted twitter twitter.new\_post}. 
	
\subsection{LSTM Encoder-Decoder}
Our baseline was a simple LSTM Encoder-Decoder model. The model was implemented and trained using the JoeyNMT framework \cite{JoeyNMT}. The encoder and decoder are bidirectional LSTMs with multiplicative attention \cite{DBLP:journals/corr/LuongPM15}. The embedding size was 16, the hidden size was 64, and the dropout factor was .10. Inputs to the encoder were right padded tokenized word ids with a fixed sequence length of 25. All words in the input and target sequence vocabularies were used. 

The model was trained on 100 epochs and evaluated every 4,000 steps. We use the adam optimizer with a learning rate of .001. Finally, we used cross-entropy as our loss function. The full configuration details for our baseline model can be found on our github repository.

\begin{table}[t]
	\begin{tabular}{l|lccc}
		& \textbf{LSTM~} & \textbf{OpenNMT} & \textbf{Transformer}  \\ 
		\hline
		Sequence         & 84.75          & 93.23            & \textbf{93.91}        \\
		Positional       & 94.39          & 97.66            & \textbf{97.94}        \\
		Event Channel  & 97.18          & 99.09            & \textbf{99.26}        \\
		Event Function & 88.97          & 94.96            & \textbf{95.47}        \\
		Action Channel   & 97.65          & 98.97            & \textbf{99.16}        \\
		Action Function  & 93.74          & 97.61            & \textbf{98.00}          
	\end{tabular}
	\caption{Seq2Seq model results on predicting Zapier recipes.} \smallskip
	\label{tab:zap_results}
\end{table}

\subsection{OpenNMT}
The second model we evaluated was the stacked RNN OpenNMT model provided in OpenNMT-py library \cite{klein-etal-2018-opennmt}. We used the default model and training script.They use a two layer RNN encoder and two layer stacked LSTM decoder. Global attention is added to the bottom. Inputs and target sequences are stored in separate files (segmented into training and validation sets) and provided to the training script. The model trained for 100,000 steps. Please see \citeauthor{klein-etal-2018-opennmt} for further model specifics. 

\subsection{Basic Transformer}
The final model we evaluated was the Transformer model introduced in \citeauthor{NIPS2017_7181} \citeyear{NIPS2017_7181}. We used the JoeyNMT framework to implement and train the Transformer model. The model consisted of six layer transfor of the architecturemer encoder and decoder with eight attention heads and used additive attention \cite{bahdanau2014neural}. The input embedding size was 512, the hidden size was 512, the feed forward size 2048 and it had dropout factor of .10. We constrained the input vocabulary to the 4000 most frequently occurring word tokens. Inputs to the encoder were right padded tokenized word ids with a fixed sequence length of 30. 

The model was trained on 100 epoch and validated every 4,000 steps. We used the noam learning rate scheduler \cite{DBLP:journals/corr/abs-1804-04235} with learning rate factor of 1 and learn rate warmup value of 4,000. We used cross-entropy for our loss function. The full configuration details for this model can be found in our github repository.

\begin{table}
	\centering
	\resizebox{\columnwidth}{!}{
		\begin{tabular}{l|ccccc}
			& \textbf{LSTM} & \textbf{OpenNMT} & \textbf{Trans.} & \textbf{Chen '16} & \textbf{Quirk '16}  \\ 
			\hline
			Sequence         & 52.35         & \textit{55.41}   & 53.06           & N/A               & N/A                 \\
			Positional       & 75.31         & \textit{76.39}   & 75.33           & N/A               & N/A                 \\
			Trigger Channel  & 75.09         & 77.44            & 74.91           & \textbf{91.60}     & 89.1                \\
			Trigger Function & 61.37         & 63.72            & 62.27           & \textbf{87.50}     & 71.00                  \\
			Action Channel   & 85.02         & 84.66            & 84.12           & \textbf{91.60}     & 89.10                \\
			Action Function  & 79.78         & 79.78            & 79.60            & \textbf{87.50}     & 71.00                 
	\end{tabular}}
	\caption{Model performance on IFTTT recipes.}
	\label{tab:ifttt_results}
\end{table}

\section{Evaluation Metrics}

In traditional language translations tasks BLEU and Rouge scores are often used as the primary evaluation metric. These scores try to account for the fact that translations may vary in word usage and syntax from the source sequences. In our task, the predicted recipes must be correct across the channels and functions. Additionally, the order of predicted sequence matters as each position in sequence corresponds to a specific recipe component.  

In addition to the recipe component accuracy scores (e.g. Trigger Channel accuracy or Event Function accuracy), we consider the overall sequence accuracy and the positional accuracy of the predicted sequences. The sequence accuracy measures if the overall sequence is correct. The prediction is assigned a score of 1 only if the predicted string exactly matches the reference string. We expect the sequence score to be lower in general as the predictions have a higher level of scrutiny.

The positional score provides partial credit to the prediction, if the tokens correctly match the reference tokens at their position in the recipe. For example, if a prediction correctly identifies 3 of the 4 recipes components, it will receive a score of .75. We provide the distribution of positional errors in Table \ref{tab:error}.

\section{Results}

We were able to successfully train all three architectures and had varying levels of success. In Table \ref{tab:ifttt_results} we see that none of the seq2seq models were able to match the reported channel/function accuracy values in \citeauthor{chen2016latent} \citeyear{chen2016latent} and \citeauthor{beltagy-quirk-2016-improved} \citeyear{beltagy-quirk-2016-improved}. Both previous works only reported aggregate channel and function accuracy values. For simplicity, we assumed those aggregate values were the same across channels and  respectively. Of the seq2seq models, OpenNMT consistently performed better than our baseline and the Transformer model. The seq2seq models did fairly well on Action Channel and Action function prediction, coming within 10 points of the \citeauthor{chen2016latent} \citeyear{chen2016latent} scores. We believe this may be related to how recipes are described in general. In both IFTTT and Zapier, users are more explicit in describing the action scenario (the "Then") part of the program than they are in describing the Triggering event. 

In contrast to IFTTT, the seq2seq models all performed strongly on the Zapier dataset (Table \ref{tab:zap_results}). Our baseline LSTM encoder-decoder that had an overall sequence accuracy or 84.75\% and strong performance on predicting each of the component recipes parts in context of the full recipe. The Transformer model performs the best is 9 points better that the baseline model in sequence accuracy. We significant improvement over Event Function (6 point increase) and Action Function (4 point increase) identification. Over we found the quality of the Zapier titles and descriptions better than those of IFTTT. Zapier titles were consistent their ability to concisely describe the recipes and the descriptions were also well written and provided useful clues about the intended channel functions.
\begin{table}
	\centering
	\resizebox{\columnwidth}{!}{
		\begin{tabular}{l|cccccc}
			\multicolumn{1}{l}{} & \multicolumn{3}{c}{\textbf{\underline{Zapier}}}             & \multicolumn{3}{c}{\textbf{\underline{IFTTT}}}               \\
			\textbf{Errors}      & \textbf{LSTM} & \textbf{ONMT} & \textbf{Trans.} & \textbf{LSTM} & \textbf{ONMT} & \textbf{Trans.}  \\ 
			\hline
			Zero                 & 84.75       & 93.23       & 9.39E-01        & 52.35       & 55.42       & 53.07           \\
			One                  & 10.54       & 5.00        & 4.73E-02        & 14.44       & 12.09       & 13.36           \\
			Two                  & 3.13        & 12.6        & 9.26E-03        & 23.10       & 21.84       & 22.92           \\
			Three                & 0.68        & 0.21        & 6.17E-04        & 2.35        & 3.97        & 2.71           \\
			Four                 & 0.91        & 0.31        & 3.70E-03        & 7.76        & 6.68        & 7.94           \\
	\end{tabular}}
	\caption{Distribution of errors across all predictions by domain. }
	\label{tab:error}
\end{table}

\section{Discussion}
Overall the seq2seq models were able to learn If-Then recipes in their entirety. The models were surprisingly robust and able to capture various linguistic variations and ambiguities found the natural language descriptions. Across both datasets, the models tended to have higher Action and Action Function scores than Trigger/Event scores. From superficial analysis, we found that users on average were more specific in describing the action then they were describing the triggering event. 

We had significant difficulty working with the IFTTT dataset. Nearly 50\% of the train and validation sets we thrown away, and nearly 90\% of the test set. Through visual and ad-hoc analysis, we would often find mistakes in both the annotated test set recipes and the scraped recipes. Given the age and volatility of the recipe urls, we were not confident that our experiment conditions matched those from previous works. We were unable to reproduce any of the prior results and therefore had difficulty doing a more through error analysis on our findings.

In Table \ref{tab:error}, we provide the distribution across all the predictions on the respective tests sets. It is interesting to note that on the IFTTT dataset, the model likely to make two errors as opposed to one, three or four. We believe there is a tight coupling between trigger channel and trigger function predictions. We hypothesize that if the model fails to predict the Trigger, it will also fail to predict the Trigger Functions.

The seq2seq model's performance on the Zapier dataset was very encouraging. As a next step we plan to investigate argument extraction on the dataset. Additionally, we are interested in explore the transfer learning potential given quality disparity between Zapier and IFTTT. Both have similiar vocabularies and recipe domains. We hypothesize a model trained on Zapier data may preform better on the IFTTT dataset than the model trained solely on IFTTT data. Finally, we are interested in investigating more complex program synthesis challenges. We believe more complex program representations can potentially be learned through seq2seq models.

\section{Conclusion}
In this paper we proposed modeling If-Then program synthesis as sequence learning task. The models attempted to learn how translate natural language descriptions into IFTTT and Zapier recipes. Three seq2seq architectures were evaluated: a baseline LSTM encoder-decoder, the OpenNMT Stacked RNN, and the Transformer model. The models were successfully trained and able to predict the full automation recipes in an end-to-end manner. Due to several challenges with IFTTT dataset, we found the seq2seq model performance to be adequate but unable to match the accuracy scores of prior work. Of the seq2seq models, the OpenNMT model performed the on the IFTTT dataset, with a sequence accuracy score of 55\%, positional accuracy of 75 \%, and overall Action Channel accuracy score of 79\%. In contrast, the seq2seq models performed strongly on the Zapier dataset. The Transformer model score the highest across all metrics. It had an sequence accuracy score of 93.91\%, positional accuracy of 97.4\%. In the future, we plan expand the scope our task to account for argument extraction and explore more complex program synthesis challenges.

\bibliographystyle{aaai}
\bibliography{dalal_submission}

\end{document}